\documentclass[runningheads,a4paper]{llncs}

\usepackage{amssymb}
\usepackage{graphicx}
\usepackage{algorithmic}
\usepackage{algorithm}
\usepackage{url}
\usepackage{mathptmx} 
\usepackage{times} 
\usepackage{amsmath} 
\usepackage{amssymb}
\usepackage{hyperref}
\usepackage[flushleft]{threeparttable}
\usepackage[T1]{fontenc}
\usepackage{color}
\usepackage{enumerate}
\usepackage{wrapfig}
\usepackage{array}
\definecolor{red}{rgb}{1.00,0.00,0.00}
\definecolor{blue}{rgb}{0.00,0.00,1.00}
\definecolor{green}{rgb}{0.4,1.00,0.0}
\definecolor{yellow}{rgb}{0.5,0.5,0.0}

\urldef{\mailsa}\path|{alfred.hofmann, ursula.barth, ingrid.haas, frank.holzwarth,|
\urldef{\mailsb}\path|anna.kramer, leonie.kunz, christine.reiss, nicole.sator,|
\urldef{\mailsc}\path|erika.siebert-cole, peter.strasser, lncs}@springer.com|    
\newcommand{\keywords}[1]{\par\addvspace\baselineskip
\noindent\keywordname\enspace\ignorespaces#1}

\begin{document}
\mainmatter  

\title{A Fast and Stable Omnidirectional Walking Engine for the Nao Humanoid Robot}
\author{Mohammadreza Kasaei\and Nuno Lau\and Artur Pereira}
\institute{	IEETA / DETI, University of Aveiro, Aveiro, 3810-193, Portugal,\\
	\email{\{mohammadreza, nunolau, artur\}@ua.pt}}

\maketitle              

\begin{abstract}
This paper proposes a framework designed to generate a closed-loop walking engine for a humanoid robot. In particular, the core of this framework is an abstract dynamics model which is composed of two masses that represent the lower and the upper body of a humanoid robot. Moreover, according to the proposed dynamics model, the low-level controller is formulated as a Linear-Quadratic-Gaussian~(LQG) controller that is able to robustly track the desired trajectories. Besides, this framework is fully parametric which allows using an optimization algorithm to find the optimum parameters. To examine the performance of the proposed framework, a set of simulation using a simulated Nao robot in the RoboCup 3D simulation environment has
been carried out. Simulation results show that the proposed framework is capable of providing fast and reliable omnidirectional walking. After optimizing the parameters using genetic algorithm~(GA), the maximum forward walking velocity that we have achieved was $80.5cm/s$.

\end{abstract}

\let\thefootnote\relax\footnotetext{This paper has been accepted for presentation at the 2019 RoboCup Symposium.} 
\vspace{-10mm}
\section{Introduction}
Developing a fast and reliable walking for a humanoid robot is a complicated subject due to dealing with a naturally unstable system. In particular, humanoid robots are known as hyper Degree of Freedom~(DOF) systems which generally have more than 20 DOFs. Humanoid robots can easily adapt to our dynamic environment because of their kinematic similarity with a human. However, developing reliable locomotion is a complex subject that absorbs the attention of researchers. Unlike wheeled robots, humanoid robots can handle real environment limitations like gaps, stairs, uneven terrain, etc. Thus, they can be used in a wide range of applications from helping elderly people to performing dangerous tasks like fire fighting.

Over the past decades, several stable walking engines have been presented and tested on real and simulated robots. Although the performance of robots in simulators are not perfectly equal with their performance in the real world, the significant advantage of using realistic simulators instead of real robot experiments is that researchers can perform much experimentation without worrying about mechanical damages on the devices of the robot (e.g., wear and tear). 

Recently, the number of researches in this field shows an increasing interest in investigating humanoid locomotion. Studies in this field can be divided into two main groups: \textit{(i)~model-based approaches}: these approaches consider a physical template model of the robot, and  a walking engine will be designed based on this model and some stability criteria; 
\textit{(ii)~model-free approaches}: researches in this group are biologically inspired and typically focus on generating some rhythmic patterns for the limbs of robot without considering any physical model of the robot. 

In this paper, we propose a fully parametric closed-loop walking framework for a simulated soccer humanoid agent. The core of this framework is an extended version of Linear Inverted Pendulum~(LIPM) which is one of the well-known walking dynamics models. LIPM considers a restricted dynamics of the COM and represents the dynamics of the robot by a first-order stable dynamics. This framework is developed and successfully tested within the RoboCup 3D simulation environment. 

The main contributions of this paper are the following: \emph{(i)} an integrated framework for humanoid walking that incorporates capabilities for walking, optimization and learning; \emph{(ii)} investigating the effect of releasing the height constraint of the COM in LIPM and examining how it can increase the walking speed and also improve the stability; \emph{(iii)} investigating how the torso motion can be used to keep the Zero Momentum Point~(ZMP) inside support polygon to provide more stable walking; \emph{(iv)} A Genetic algorithm~(GA) is used to find the optimum value for the walking parameters and the optimized walking engine is tested within the RoboCup 3D simulation environment.

The remainder of this paper is organized as follows: Section~\ref{sec:related_work} gives an overview of related work. In Section~\ref{sec:Dynamics}, firstly, the gait stability criterion is explained, then, an overview of the LIPM is presented, afterward, the effects of releasing the height constraint of the COM and considering the mass of torso are discussed. Section~\ref{sec:lowlevelcontrol} explains the structure of our robust low-level controller. Then, the overall architecture of our framework is presented and explained in Section~\ref{sec:walkengine}. Simulation scenarios, optimization method and the results are presented in Section~\ref{sec:simulation}. Finally, conclusions and future research are presented in Section~\ref{sec:conclusion}.

\section{Related Work}
\label {sec:related_work}
Successful approaches exist in both model-based and model-free biped locomotion groups and each group has its advantages and disadvantages. For instance, although model-free approaches avoid dealing with dynamics model of the system, their structure is composed of several oscillators which have many parameters that should be tuned. Besides, adopting sensory feedback to all the oscillators is a complicated subject. Indeed, tuning the parameters for performing a particular motion is typically based on optimization algorithms which are iterative procedures and need to a large number of tests. In the remainder of this section, we briefly review some approaches of both groups and focus on those which are more related to this paper.

One of the popular ways to develop a model-based walking is using a dynamics model of the robot and the concept of ZMP. ZMP is a point on the ground plane where the ground reaction force acts to compensate gravity and inertia~\cite{vukobratovic1970stability}. The fundamental idea of using ZMP as a criterion is that keeping the ZMP within the support polygon guarantee the stability of the robot. Based on this idea, generally, a predefined ZMP trajectory is used to generate a feed-forward walking. Given that the feed-forward approaches are not stable enough due to their sensitivity to the accuracy of the dynamics model, a feedback controller should be used to keep the ZMP inside the supporting polygon. Kajita et al.~\cite{kajita2003biped} proposed a walking pattern generator based on LIPM and preview control of ZMP. They conducted a simulation to show the performance of their approach. In this simulation, a simulated robot~(HRP-2P) should walk on spiral stairs. They specified the foot placement in advance and based on that the trajectories of the COM were generated. Simulation results showed that the robot could successfully walk on the spiral stairs. Nowadays, their approach is widely used and has been successfully implemented on different humanoid platforms. Although their method has several benefits such as simplicity in implementation, it also has some drawbacks like needing to specify future steps in advance and also requiring more computational power in comparison with simple methods because its use of an optimization method. Another variant of this method has been proposed which used the analytical solution of the ZMP equation to generate the COM trajectories~\cite{sugihara2005fast},~\cite{harada2006analytical},~\cite{morisawa2007experimentation}. This method is sufficient for robots with constrained computation resources. 

LIPM assumes motion of the upper body parts are negligible and do not have effects in the overall dynamics of the robot. Although considering this assumption simplifies the walking control problem, it is not an appropriate assumption because the upper body of a humanoid robot has several DOFs that their motions generate momentum around COM. Moreover, controlling this momentum not only causes significant improvements in stability of the robot but also can be used as a push recovery strategy. Therefore, several extended versions of LIPM have been proposed to improve the accuracy of the dynamics model of the robot. Napoleon et al.~\cite{nakaura2002balance} used two masses inverted pendulum to represent the lower and the upper body of a humanoid robot in their dynamics model and proposed a ZMP feedback controller based on this model. Later, Komura et al.~\cite{komura2005feedback} and also Pratt et al.~\cite{pratt2006capture} considered angular momentum around COM in their dynamics model and proposed an extended version of LIPM. They conducted several simulations and showed that using their model, the robot was not only able to generate walking but also could regain its stability after applying pushes. Recently, Kasaei et al.~\cite{kasaei2018optimal} extended this model by considering the mass of stance leg to improve the accuracy of the controller. Using this model, they proposed an optimal closed-loop walking engine and showed the performance of their system using walking and push recovery simulation scenarios.

Several model free approaches as possible alternative to generate humanoid locomotion have been proposed. The approaches based on the concept of Central Pattern Generators~(CPG) are one of the important categories in this group. Picado et al.~\cite{picado2009automatic} used some Partial Fourier Series~(PFS) oscillators to generate a forward walk for a simulated humanoid robot and used GA to optimize the parameters of the oscillators. They showed the optimized version of their approach was able to achieve a fast forward walking~($51~cm/s$). Later, Shahriar et al.~\cite{asta2011nature} extended their approach and developed a walking engine which not only was able to generate faster forward walking but also able to generate stable sidewalk. It should be noted that several hybrid approaches have been  proposed~\cite{kasaei2017hybrid},~\cite{or2010hybrid},~\cite{massah2013hybrid} which used the ZMP concept to modify the outputs of the oscillators, consequently providing more stable walking. Some of these approaches, try to find an appropriate way to adapt sensory feedback based on classical control approaches. In some others approaches, using learning algorithms, robots learn how to modify the output of each oscillator to track the reference trajectories and compensate the errors. For instance, in~\cite{gay2013learning}, a framework based on neural network has been developed to learn a model-free feedback controller to balance control of a quadruped robot walking on rough terrain. 

In most of the proposed walking engines described above, the height of COM was considered to be fixed, moreover, the effects of the motion of the torso were not considered. In the rest of this paper, we propose a closed-loop model based walking framework and show how the motion of torso and changing the height of COM can improve the walking performance. 

\section{Gait Stability and Dynamics Model}
\label{sec:Dynamics}
Our proposed framework stays on the model-based group and the core of this framework is an abstract dynamics model of the robot. In this section, firstly, we briefly describe the ZMP as our main criterion for gait stability and then present an overview of LIPM and investigate the effects of changing the height of COM as well as moving the torso.
\subsection{ZMP and Gait Stability }
Several criteria for analyzing the stability of the robot have been proposed~\cite{goswami1999postural,vukobratovic1970stability}. According to the nature of normal human gait which is composed of 80\% of single support and 20\% of double support~\cite{winter1990control}, most of the proposed criteria focus on the single support phase of walking. In this phase, one foot is in contact with the ground and the other one swings forward. In this work, the concept of ZMP is used as our main criterion to analyze the stability of the robot and is defined using the following equation:
\begin{equation}
p_x = \frac{\sum_{k=1}^{n} m_k x_k (\ddot{z}_k + g) - \sum_{k=1}^{n} m_k z_k \ddot{x}_k  }{\sum_{k=1}^{n} m_k (\ddot{z}_k + g)} \quad , 
\label{eq:zmp}
\end{equation}
\noindent
where $n$ is the number of masses that are considered in the dynamics model, $m_k$ is the mass of each part, $(x_k,\dot{x}_k)$,~$(\ddot{z}_k,\ddot{z}_k)$ represent the horizontal and vertical position and acceleration of each mass, respectively. 

\subsection {Dynamics Model}
LIPM is one of the well-known dynamics model which is used to generate and analyze the locomotion of robots. This model abstracts the dynamics of the robot using a single mass that is connected to the ground via a massless rod. This model considers an assumption that the mass is restricted to move along a defined horizontal plane. In this model, the motion equation in sagittal and
frontal planes are equivalent and independent. Therefore, we derive the equation in the sagittal plane. According to the Equation~\ref{eq:zmp} and based on LIPM assumptions, the overall dynamics of the robot can be represented using a first-order stable dynamics as follows:
\begin{equation}
\ddot{x}_c = \omega^2 ( x_c - p_x) \quad ,
\label{eq:lipm}
\end{equation}
\noindent
where  $\omega = \sqrt{\frac{g+\ddot{z}}{z}}$ represents the natural frequency of the pendulum, $x_c$ and $p_x$ are the positions of COM and ZMP, respectively. According to the assumptions of LIPM, knee joints have to be crouched to keep the COM at a constant height which consumes more energy during walking. In order to generate more energy efficient and also more human-like locomotion, a sinusoidal trajectory is assigned to the vertical motion of COM. The vertical trajectory of COM in this model is as follows:
\begin{equation}
z_c = z_0 +  A_z\cos(\frac{2\pi}{StepTime} t + \phi)\quad ,
\label{eq:verticalCOM}
\end{equation}
\noindent
where $Z_0$ is the initial height of COM, $A_z$, $\phi$ represent the amplitude and the phase shift of vertical sinusoidal motion of the COM, respectively. It should be mentioned that these parameters can be adjusted using sensory feedback to deal with environmental perturbations. However, the current version of dynamics model is good enough to generate a fast walking but it is not suitable for generating a very fast walking. During a very fast walking, COM accelerates forward and ZMP moves behind the center of gravity~(COG), based on this dynamics model, COM should be decelerated to keep the ZMP inside the support polygon. This deceleration causes the ZMP to move to the boundary of the support
foot and consequently, the robot starts to roll over. One of the possible approaches to regain ZMP inside the support foot is using the torso movement for compensating the ZMP error and prevent falling.
\begin{figure}[!t]
	\label {fig:Models}
	\centering
		\begin{tabular}	{c c c}			
			\includegraphics[width=0.28\textwidth, trim= 10.5cm 2.5cm 10cm 2cm,clip] {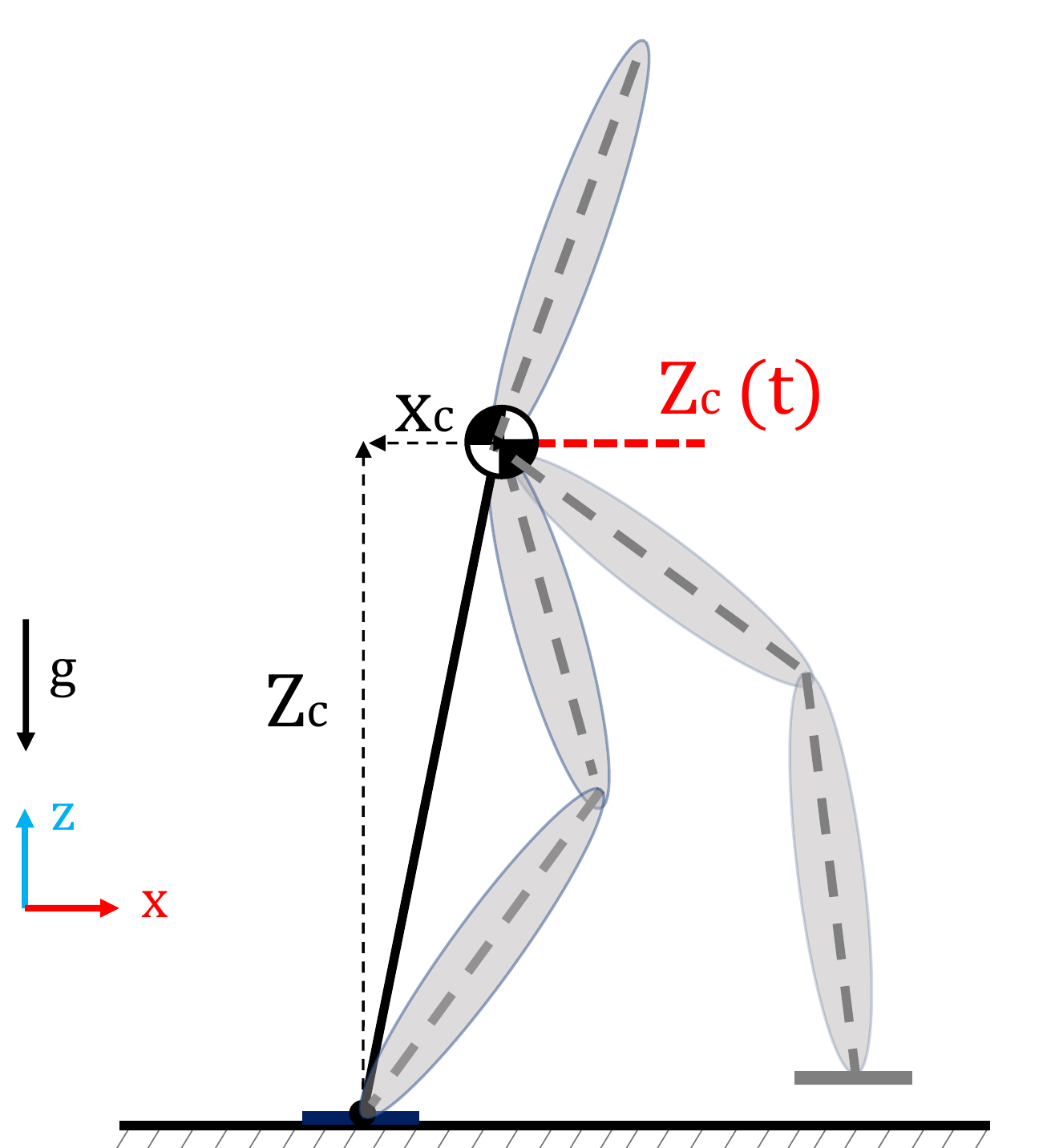} &
			\includegraphics[ width=0.28\textwidth,trim= 10.5cm 2.5cm 10cm 2cm,clip] {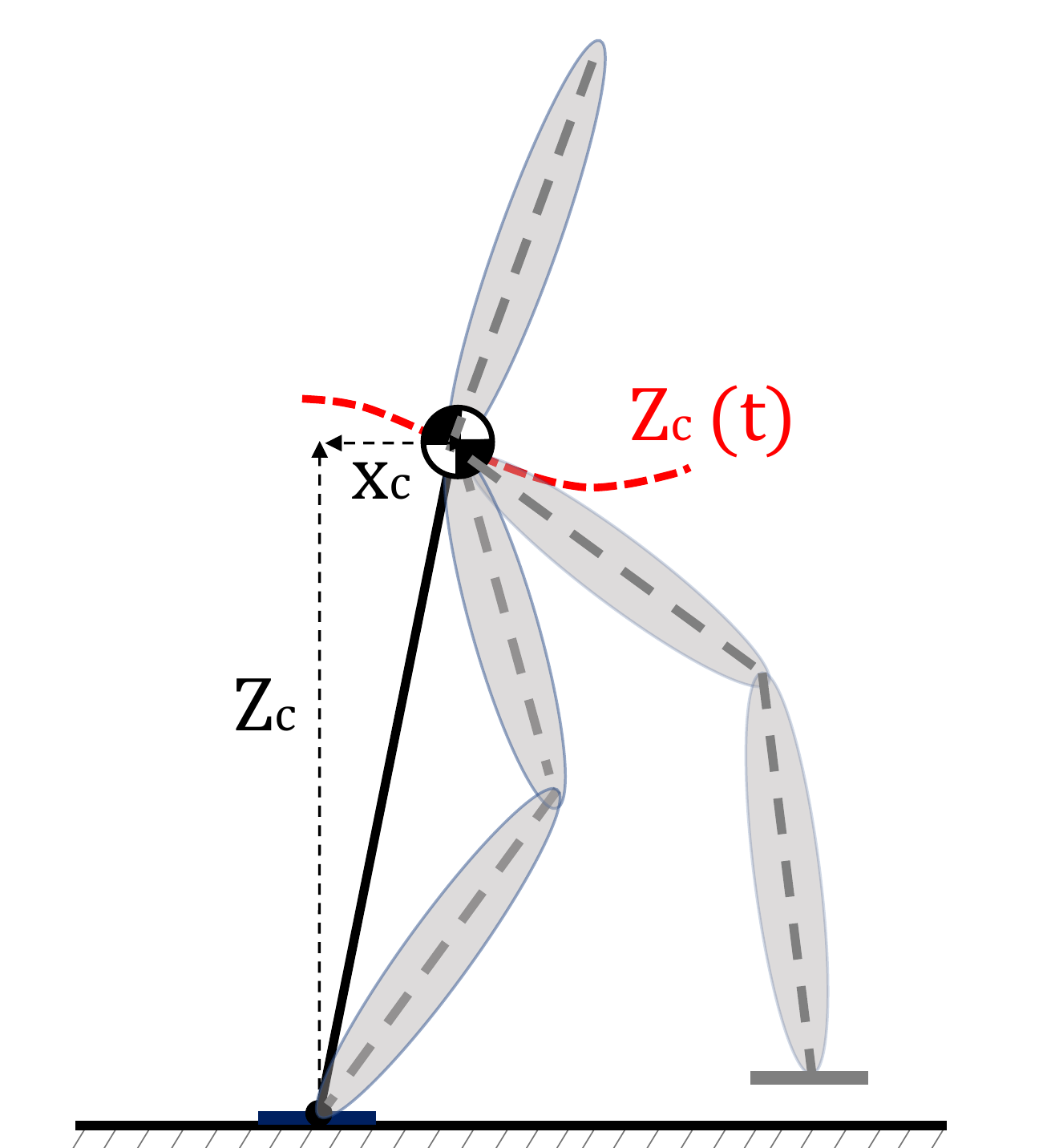}&
			\includegraphics[width=0.28\textwidth,trim= 10.5cm 2.5cm 10cm 2cm,clip] {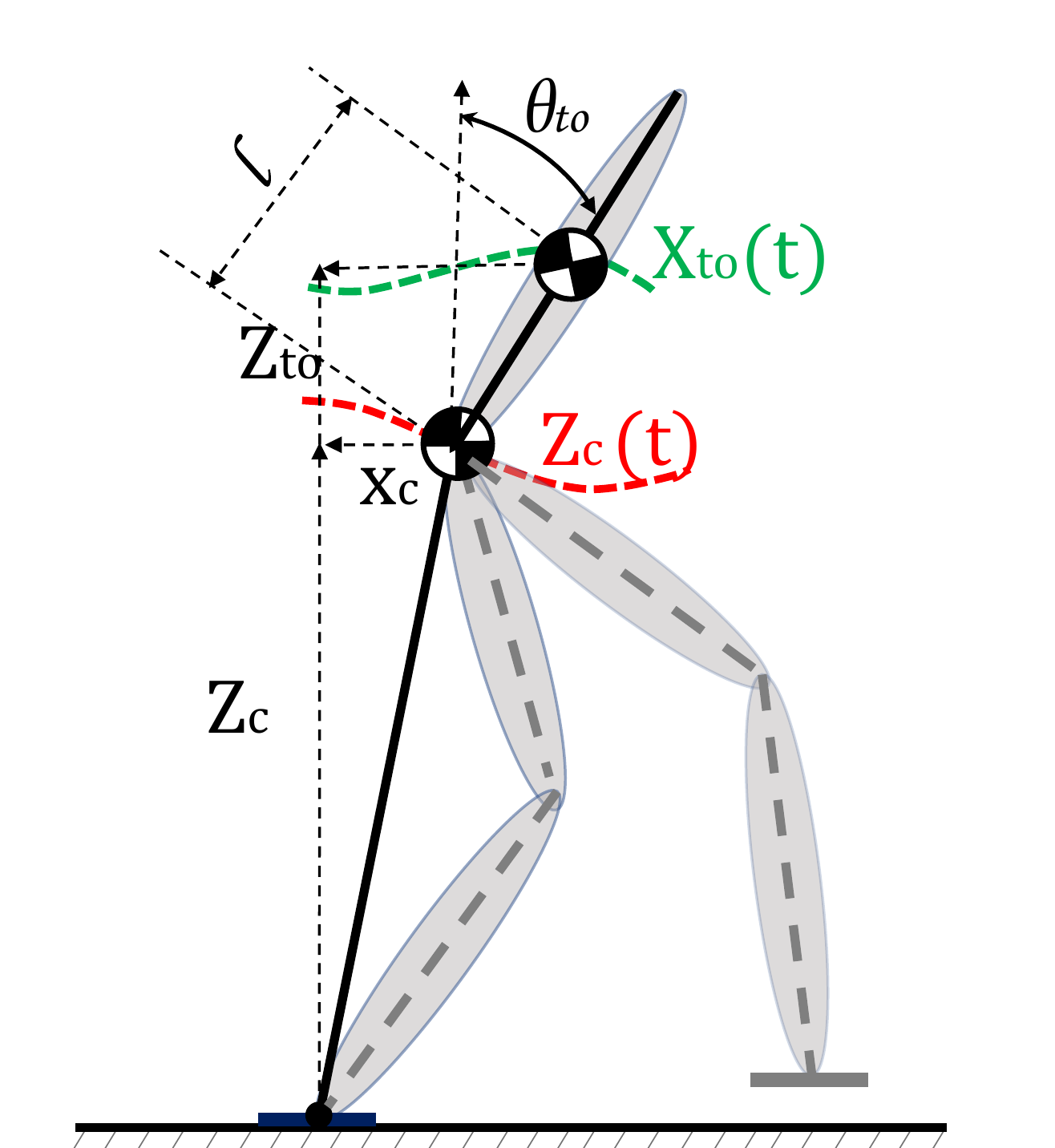}\\
			(a) & (b) & (c)
		\end{tabular}
	\vspace{-0mm}
	\caption{ Schematics of the dynamics models: (a) LIPM; (b) LIPM with vertical motion of COM; (c) LIPM with vertical motion of COM and torso motion.}
	\vspace{-5mm}
	\label{fig:DynamicsModel}
\end{figure}
 Although it is possible to modify the dynamics model by adding another mass as a torso, this modification changes our linear dynamics into a nonlinear dynamics model which does not have an analytical solution generally and should be solved numerically. According to the biomechanical analysis of human walking and running, torso moves in a sinusoidal form and the amplitude of its movement depends on some parameters like walking speed, the amplitude of changing the height of COM, etc. To consider the effect of torso motion based on biomechanical analysis, a mass with a sinusoidal movement is added to our dynamics model. Therefore, the motion equation of this model can be obtained using Equation~(\ref{eq:zmp}) as follows:
\begin{equation}
\begin{aligned}
\ddot{x}_c =  \mu ( x_c+\frac{\alpha l}{1+\alpha}&\theta_{to} - p_x)- \frac{\alpha\beta l}{1+\alpha\beta}\ddot{\theta}_{to} \\
	\alpha = \frac{m_{to}}{m_c}, \quad \beta = \frac{z_{to}}{z_{c}}, \quad \mu =& \frac{1+\alpha}{1+\alpha\beta}\omega^2, \quad x_{to} = x_c+l\theta_{to}
\end{aligned}
\label{eq:lipm_new}
\end{equation}
\noindent
where $x_{to}$ is the position of torso, $l$, $\theta_{to}$ represent the torso length and torso angle,
$m_{to}$, $m_{c}$, $z_{to}$, $z_{c}$ are the masses and heights of torso and lower body, respectively. Equation~\ref{eq:lipm_new} can be represented as a linear state space system as follows:
\begin{equation}
\label{eq:statespace_alpha}
\begin{aligned}
\dot{x} = Ax&+Bu\\
\begin{bmatrix} \dot{x}_c\\ \ddot{x}_c \\ \dot{\theta}_{to} \\ \ddot{\theta}_{to} \end{bmatrix}
	=
		\begin{bmatrix} 
		0 & 1 & 0 & 0  \\ 
		\mu & 0 & \frac{\mu\alpha l}{1+\alpha} & 0  \\
		0 & 0 & 0 & 1\\
		0 & 0 & 0 & 0				  
	\end{bmatrix}
	&\begin{bmatrix} x_c\\ \dot{x}_c \\ \theta_{to} \\ \dot{\theta}_{to}\end{bmatrix}
	+
	\begin{bmatrix} 
		0 & 0  \\
		-\mu &\frac{-\alpha\beta l}{1+\alpha\beta}\\
		0 &0\\
		 0 &1
	\end{bmatrix}
	\begin{bmatrix} 
		p_x \\ \ddot{\theta}_{to}  
	\end{bmatrix}.
\end{aligned}	
\end{equation}
In the next section of this paper, we will explain how this state space system can be used to design an optimal low-level controller to generate a very fast and stable walking.

\section{Low Level Controller}
\label{sec:lowlevelcontrol}
\begin{wrapfigure}{r}{50mm}
\vspace{-20mm}
	\center
	\includegraphics[ scale = 0.18,trim= 4.5cm 6.8cm 4.5cm 6.3cm,clip]{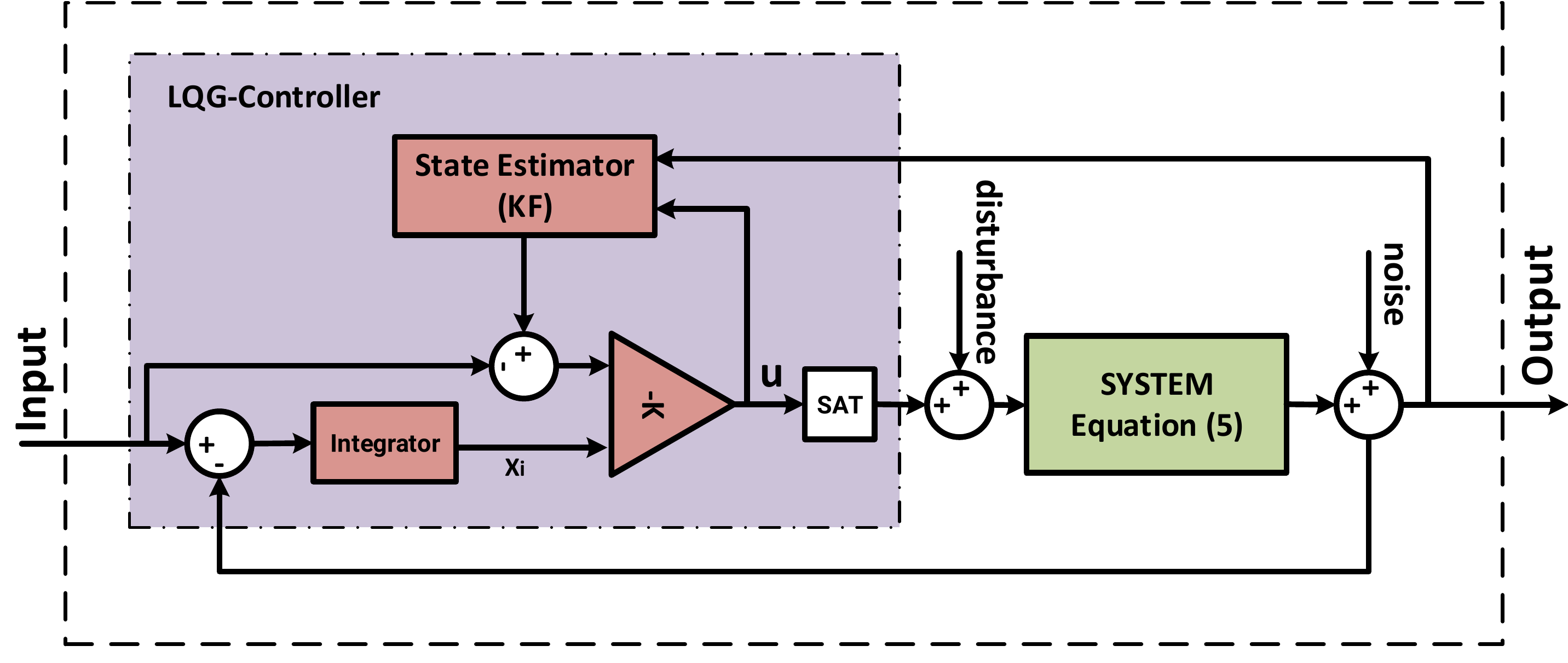}
	\caption{Block diagram of the proposed low-level controller.}
	\label{fig:ControllerArchitecture}       
	\vspace{+15mm}
\end{wrapfigure} 
In this section, an optimal state-feedback controller will be designed based on the obtained state space system in the previous section. This controller is a Linear-Quadratic-Gaussian~(LQG) controller which is robust against process disturbances and also measurement noise. The overall architecture of this controller is depicted in figure~\ref{fig:ControllerArchitecture}. As is shown in this figure, the controller is composed of an integrator for eliminating steady-state error and also two other fundamental modules which are the state estimator and the optimal controller gain that will be described in the following of this section. 
\subsection{State Estimator}
\begin{wrapfigure}{r}{50mm}
	\vspace{-45mm}
	\center
	\begin{tabular}	{c}		
		\includegraphics[ scale = 0.11]{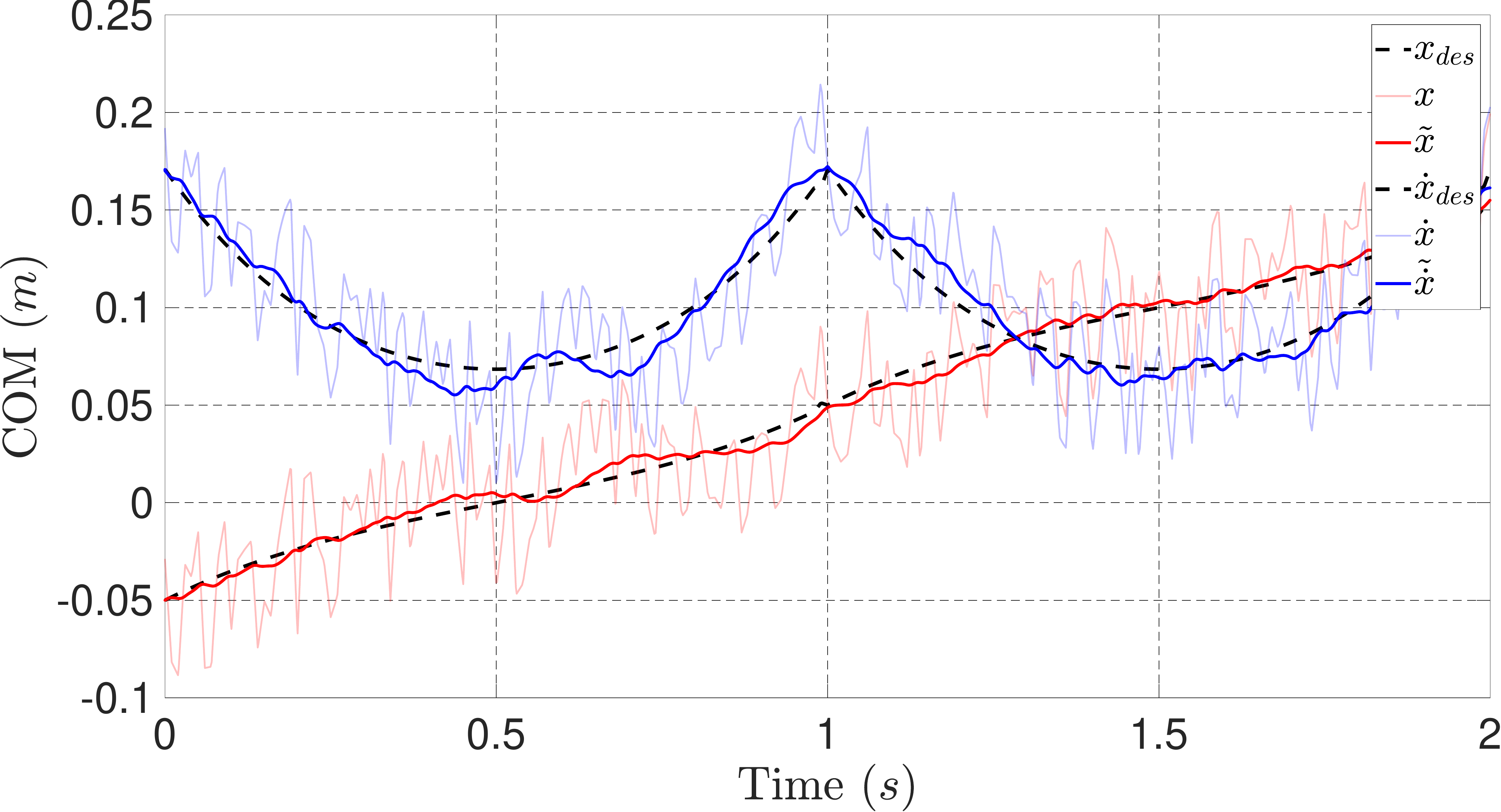}\\
		\includegraphics[ scale = 0.11]{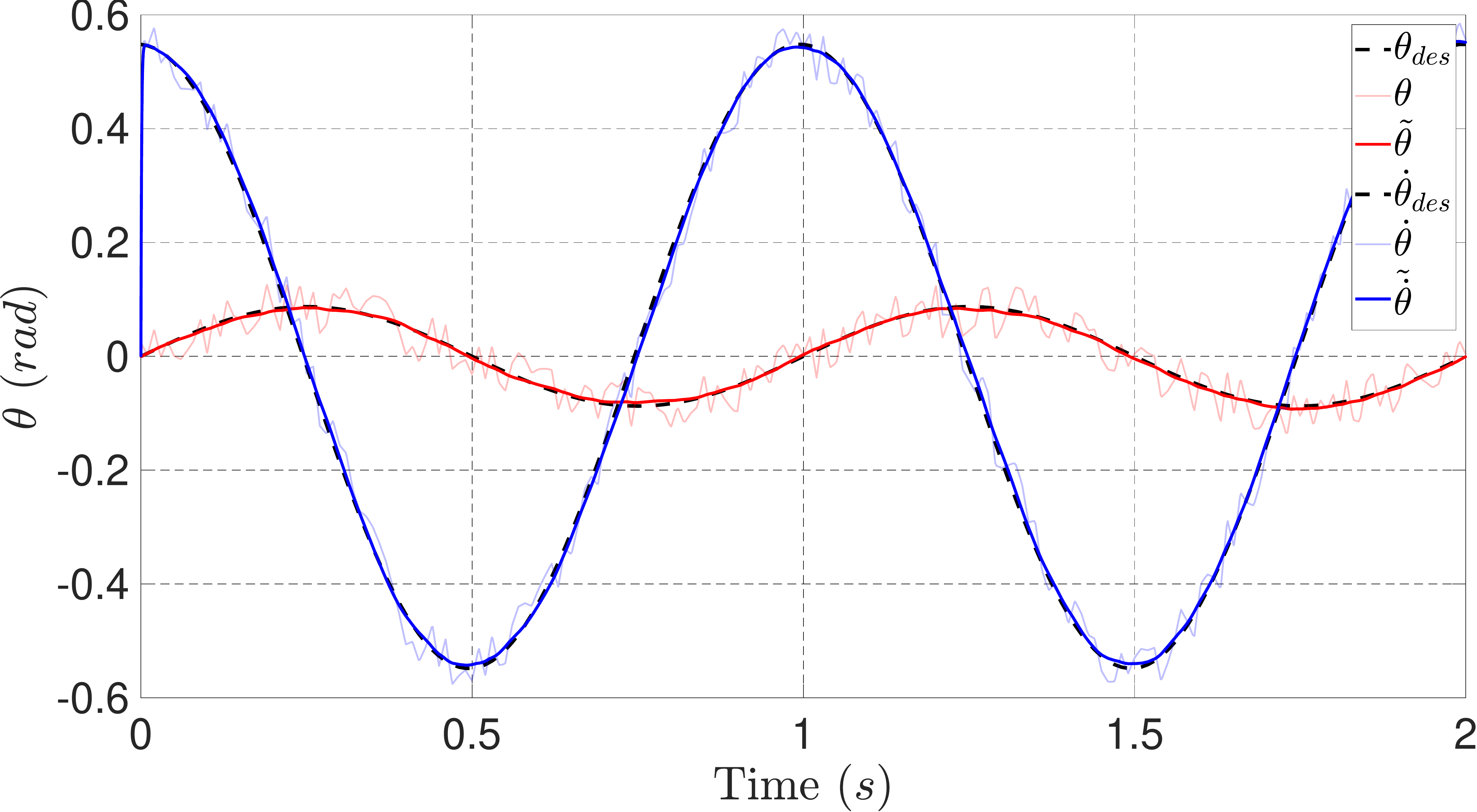}		
	\end{tabular}	
	\caption{Simulation results of the examining the state estimator performance .}
	\label{fig:Kalman}       
	\vspace{-8mm}
\end{wrapfigure} 
Typically measurements are affected by noise that occur because of modeling errors and sensors noise. In particular, to design a controller which could be able to robustly track the reference trajectories in presence of uncertainties, firstly, the states of the system should be estimated based on the control inputs and the observations. In this paper, we assume that the states of the system are observable but the observations are noisy. According to this assumption and also linearity of the dynamics system~(Equation~\ref{eq:statespace_alpha}), a Kalman Filter~(KF) was used to estimate the states of the system. Indeed, KF is a recursive filter that can estimate the states of a linear dynamics system in the presence of noise. To examine the performance of the KF, we carried out a simulation by modeling the observations as a stochastic process by applying two independent Gaussian noises to the measured states. The simulation results are depicted in figure~\ref{fig:Kalman}. As is shown, KF was be able to estimate the states of the system. 

\subsection{Optimal Controller Gain}
\begin{figure}[!t]
	\centering
		\begin{tabular}	{c  c}			
			\includegraphics[width=0.4\textwidth] {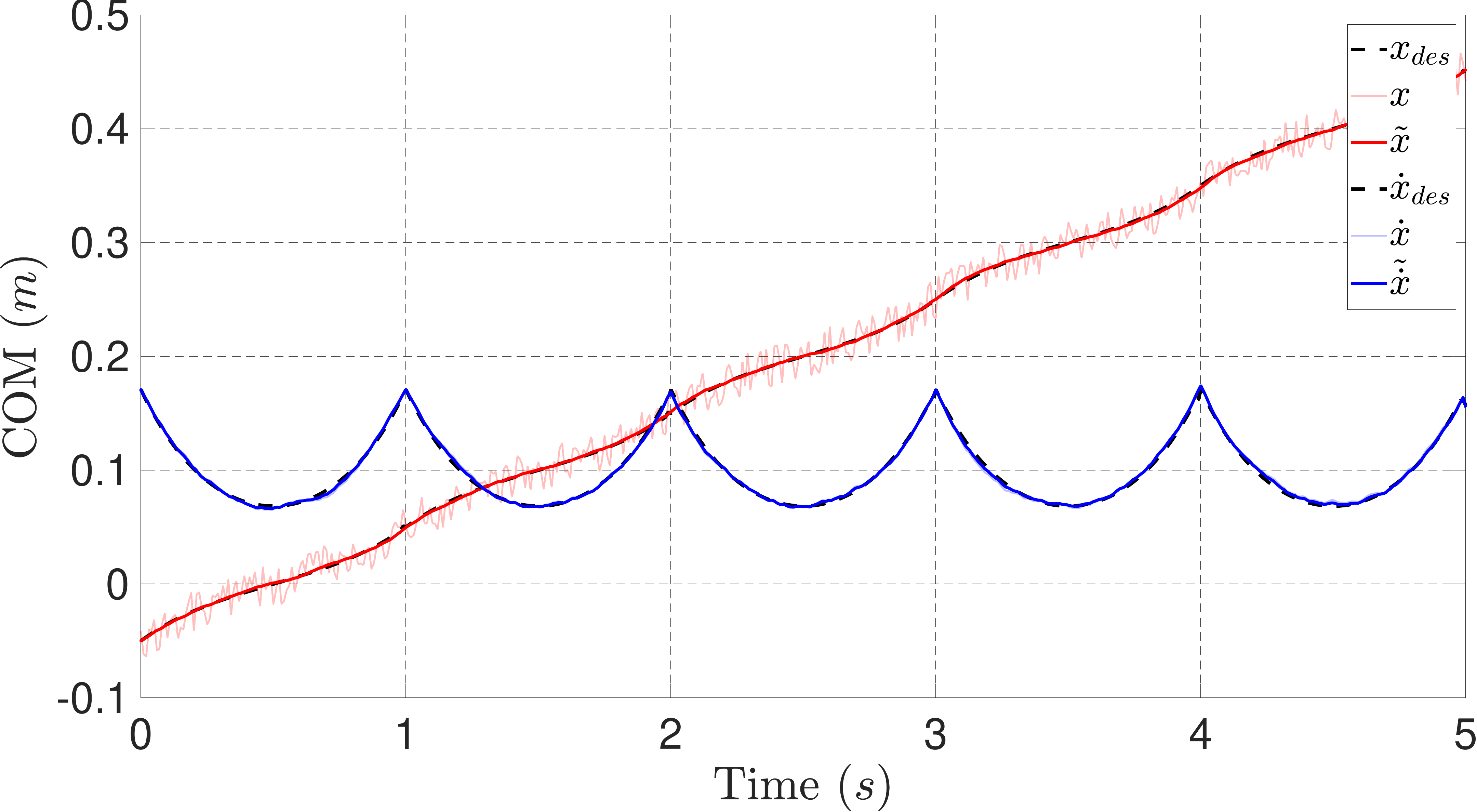} &
			\includegraphics[width=0.4\textwidth] {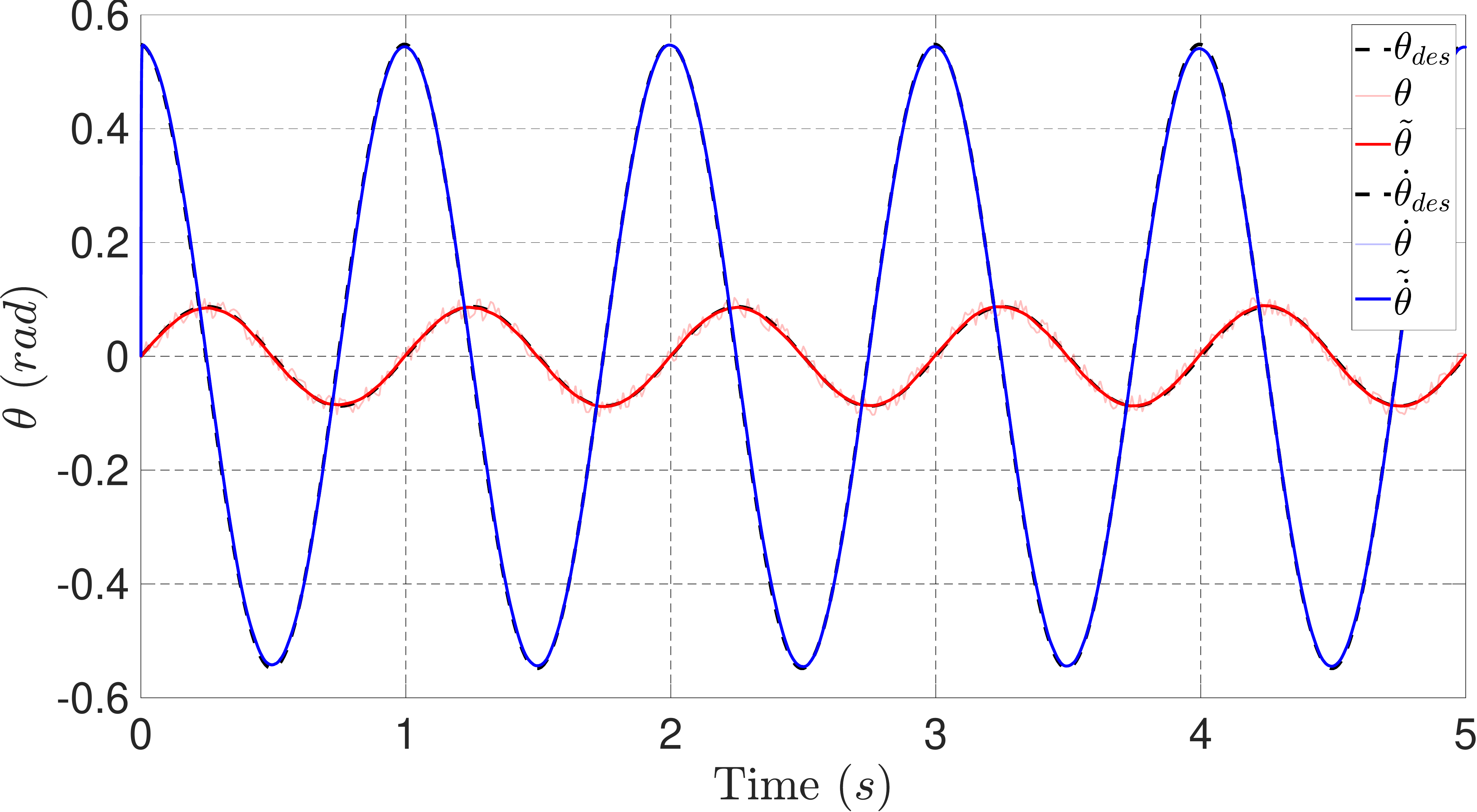}
		\end{tabular}
	\vspace{-2mm}
	\caption{Simulation results of examining the controller performance in presence of noises. }
	\vspace{-0mm}
	\label {fig:controller_test}
\end{figure}
Based on the estimation of the states and also the integration of error which are available in each control cycle, we formulate the control problem as optimal control. Indeed, this controller finds a control law for our system based on a cost function which is a function of state and control variables. The optimal control law for the tracking is designed as follows:
\begin{equation}
u = -K
\begin{bmatrix}
\tilde{x} - x_{des}\\
x_i
\end{bmatrix} \quad ,
\end{equation}
\noindent
where $\tilde{x}$, $x_{des}$ are the estimated states and the desired states, respectively. $x_i$ represents the integration of error, $K$ is the optimal gain of the controller which is designed to minimize the following cost function:
\begin{equation}
J(u) = \int_{0}^{\infty} \{ z^\intercal Q z + u^\intercal R u \} dt \quad ,
\end{equation}
\noindent
where $z = [\tilde{x} \quad x_i]^\intercal $, $Q$ and $R$ are a trade-off between
tracking performance and cost of control effort. A straightforward solution exists for finding the $K$ based on the solution of a differential equation which is called the Riccati Differential Equation (RDE). The performance of the controller is sensitive to the choice of $Q$ and $R$.  It should be noted that they are positive-semidefinite and positive-definite and selected using some trial and error. To examine the performance of the low-level controller, a simulation has been carried out. In this simulation, the controller should track a reference trajectory in the presence of noise. In this simulation, in order to simulate a noisy situation, independent zero-mean Gaussian noises are added to the measurements. The simulation results are depicted in figure~\ref{fig:controller_test}. The results show that the controller is able to robustly track the reference even in a noisy situation. In the next section, we will explain how this low-level controller can be used to generate stable walking.
\section{Walk Engine}
\label{sec:walkengine}
\begin{wrapfigure}{r}{48mm}
	\vspace{-12mm}
	\center
	\includegraphics[ scale = 0.2,trim= 4cm 1.8cm 3.5cm 2.2cm,clip]{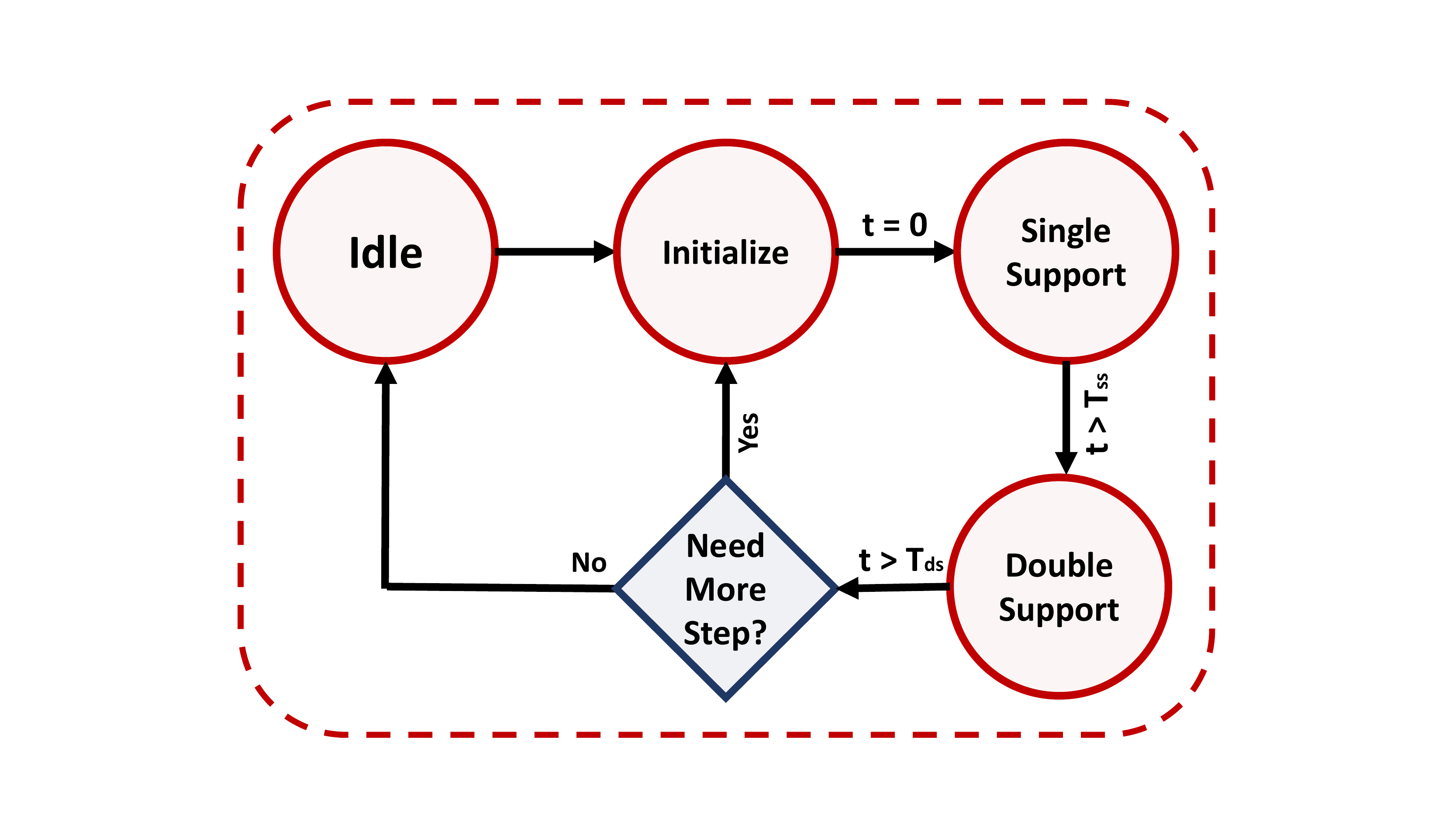}		
\vspace{-8mm}
	\caption{Walk engine state machine .}
	\label{fig:statemachine}       
	\vspace{-5mm}
\end{wrapfigure} 
In this section, based on the presented dynamics model and the low-level controller in previous sections, a walking engine will be designed. Walking is a periodic motion which is composed of four phases: Idle, Initialize, Single Support, Double Support. In the Idle phase, the robot is loading the initial parameters from its database and standing in place and waiting for a walking command. The walking phase will be changed to the initialize phase once a walking command is received. In this phase, the robot moves its COM from initial position to the first single support foot and ready to start walking procedure. During single and double support phase walking trajectories will be generated and commanded to the low-level controller to generate walking motion. This walking engine can be modeled by a state machine which has four distinct states. In this state machine, each state has a specific duration and a timer is used to trigger state transitions. The architecture of this state machine is depicted in figure~\ref{fig:statemachine}. 

\section{Reference Trajectories Planner}
Generally, a walking reference trajectories planner is composed of five sub planners which are connected together in a hierarchical manner. The first level of this hierarchy is foot step planner which generates a set of foot positions based on given step information, terrain information and some predefined constraints (e.g., maximum and minimum of the step length, step width, the distance between feet, etc.). The second planner is the ZMP planner that uses the planned foot step information to generate the ZMP reference trajectories. In our target framework, our ZMP reference planner is formulated as follows:
\begin{equation}
r_{zmp}= 
\begin{cases}
\begin{cases}
f_{i,x} \\
f_{i,y} \qquad\qquad\qquad\qquad\qquad\qquad 0 \leq t < T_{ss} \\
\end{cases} \\
\begin{cases}
f_{i,x}+ \frac{L_{sx} \times (t-T_{ss})}{T_{ds}}   \\
f_{i,y}+\frac{L_{sy}\times (t-T_{ss})}{T_{ds}} \qquad\qquad\qquad T_{ss} \leq t < T_{ds} \\
\end{cases} 
\end{cases} ,
\label{eq:zmpEquation}
\end{equation}
\noindent
where $f_i = [f_{i,x} \quad f_{i,y}]$ are the planned footsteps on a 2D surface ($i \in \mathbb{N}$), $L_{sx}$ and $L_{sy}$ represent step length and step width, $T_{ss}$, $T_{ds}$ are the single support duration and double support duration, respectively. $t$ is the time which will be reset at the end of each step ($t\geq T_{ss}+T_{ds}$). The third planner is the swing leg planner which generates the swing leg trajectory using a cubic spline function. This planner uses three control points that are the position of swing leg at the beginning of the step, the next footstep position and a point between them with a predefined height~($Z_{swing}$). The fourth planner is the global sinusoidal planner which generates three sinusoidal trajectories for the height of COM, the torso angles and the arm positions. The fifth planner is the COM planner which uses the planned ZMP trajectories and the analytical solution of the LIPM to plan the COM trajectories. This planner is formulated as follows:
\begin{equation}
\label{eq:com_traj_x0xf}
	x(t) = r_{{zmp}_x} + \frac{ (r_{{zmp}_x}-x_f) \sinh\bigl(\omega(t - t_0)\bigl)+ (x_0 - r_{{zmp}_x}) \sinh\bigl(\omega(t - t_f)\bigl)}{\sinh(\omega(t_0 - t_f))},
\end{equation}
\noindent
where $r_{{zmp}_x}$ represents the current ZMP position, $t_0$, $t_f$, $x_0$, $x_f$ are the times and corresponding positions of the COM at the beginning and at the end of a step, respectively. In this work, $T_{ds}$ is considered to be zero, it means ZMP transits to the next step at the end of each step instantaneously~\cite{kasaei2018optimal}. Moreover, $x_f$ is assumed to be in the middle of current support foot and next support foot ($\frac{f_{i} + f_{i+1}}{2}$).

\section{Result}
\label{sec:simulation}
\begin{figure}[!t]
	\centering
	\includegraphics[scale=0.29, trim= 5cm 8.5cm 5cm 8cm,clip] {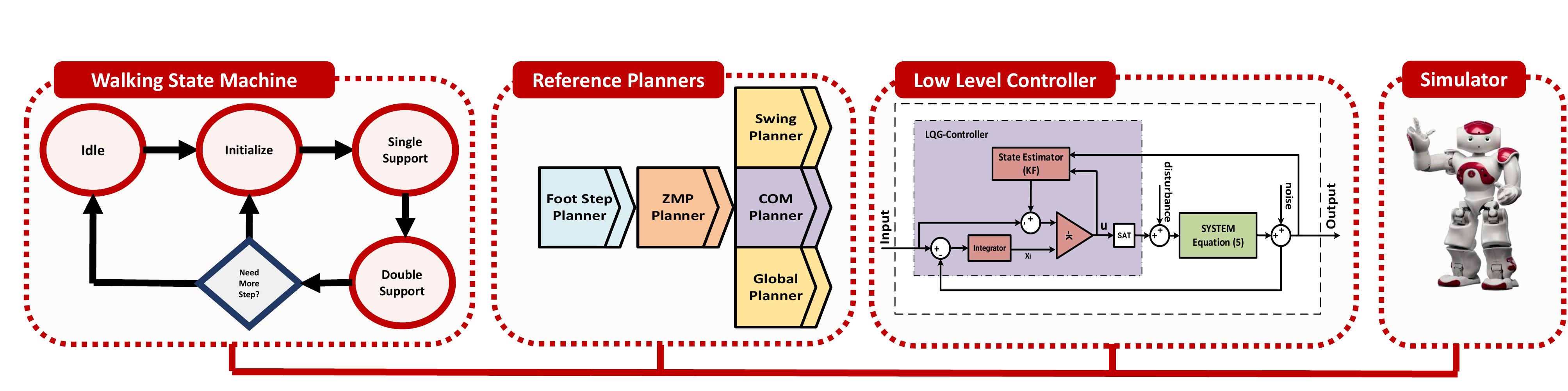} 
	\caption{ Overall architecture of the proposed walking engine.}
	\vspace{-5mm}
	\label{fig:OveralArch}
\end{figure}

To validate the performance of our proposed walking engine, a simulation scenario is designed which focused on omnidirectional walking. This simulation scenario has been set up using the RoboCup 3D simulation league simulator which is based on SimSpark~\footnote{http://simspark.sourceforge.net/} multi-agent simulator. This simulator simulates rigid body dynamics using the Open Dynamics Engine~(ODE) and provides a realistic simulation. In this simulator, the physics simulation is updated every $0.02~s$. The overall architecture of this simulation setup is depicted in figure~\ref{fig:OveralArch}.

\begin{figure}[!h]
	\centering
	\begin{tabular}	{c c c c }	
	\includegraphics[width=0.23\textwidth, trim= 18cm 11cm 20cm 10cm,clip] {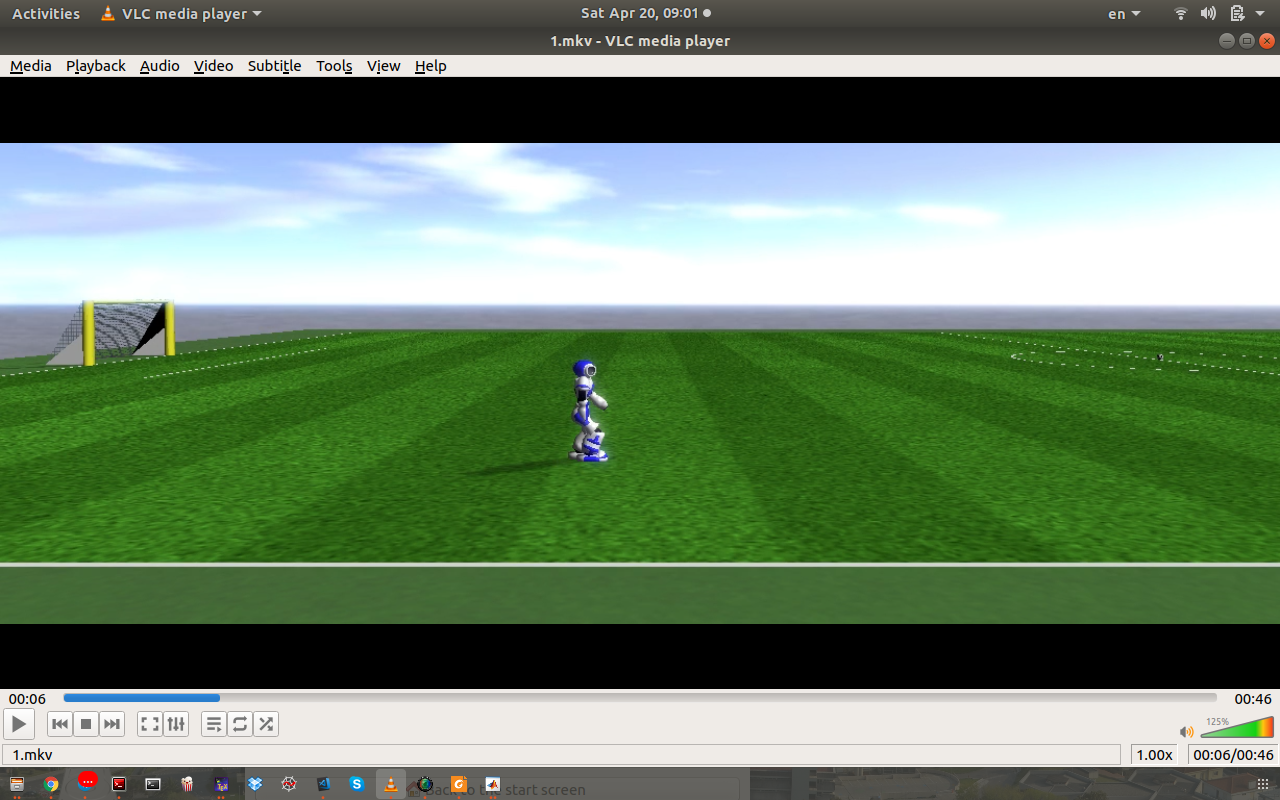}&
	\includegraphics[width=0.23\textwidth, trim= 18cm 11cm 20cm 10cm,clip] {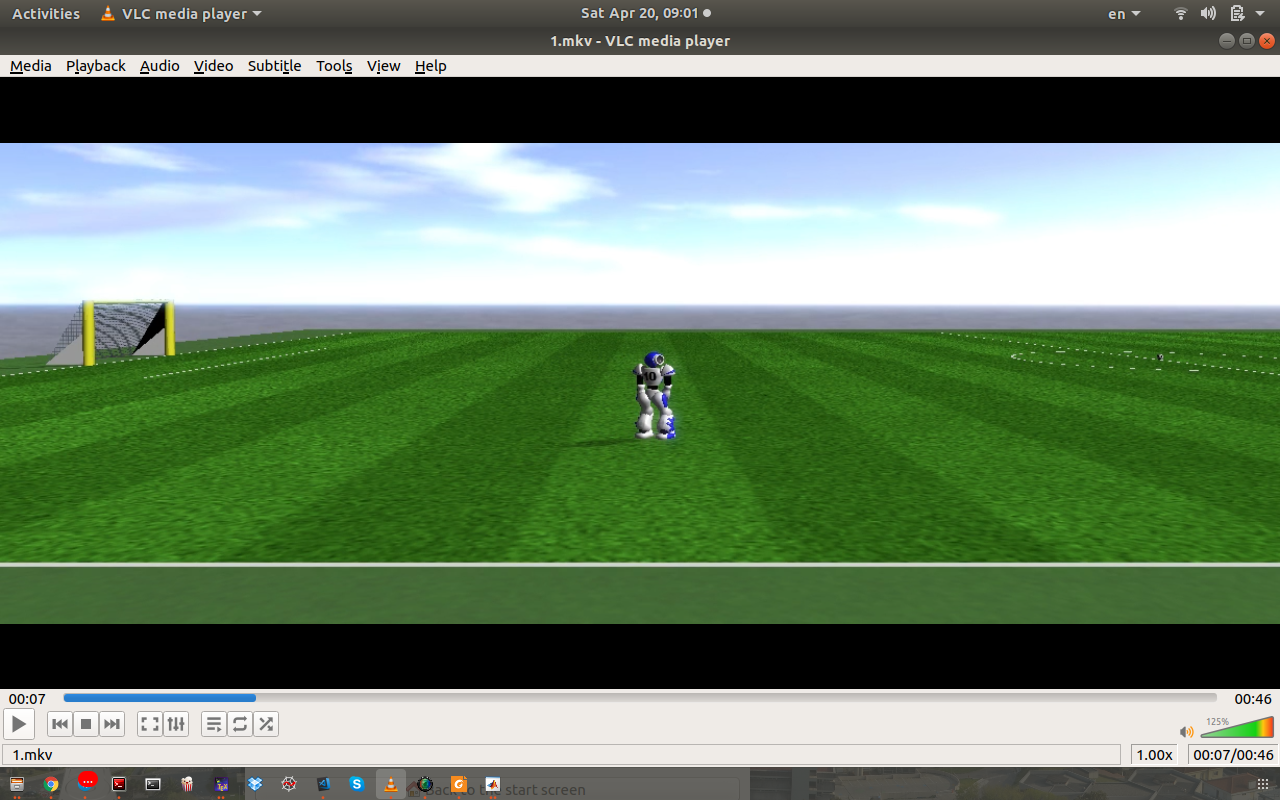}&
	\includegraphics[width=0.23\textwidth, trim= 10cm 11cm 28cm 10cm,clip] {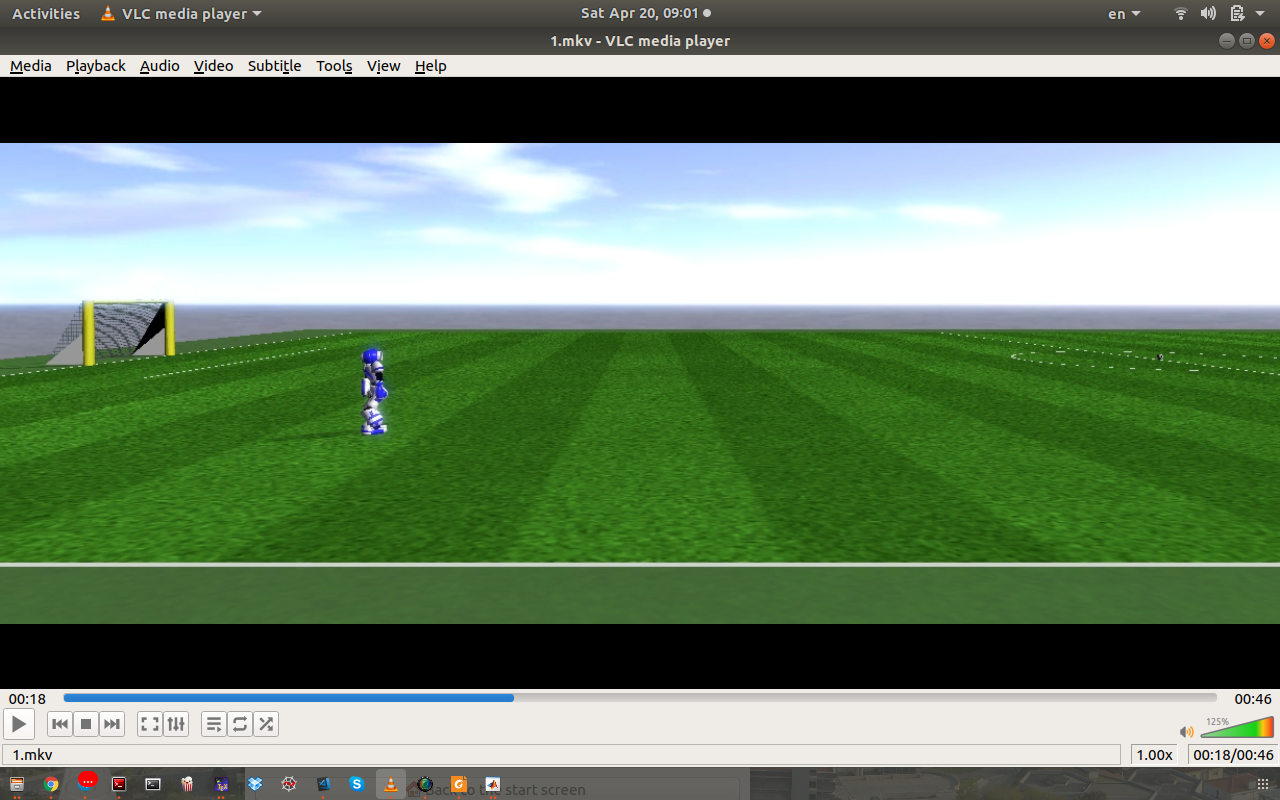}&
	\includegraphics[width=0.23\textwidth, trim= 10cm 11cm 28cm 10cm,clip] {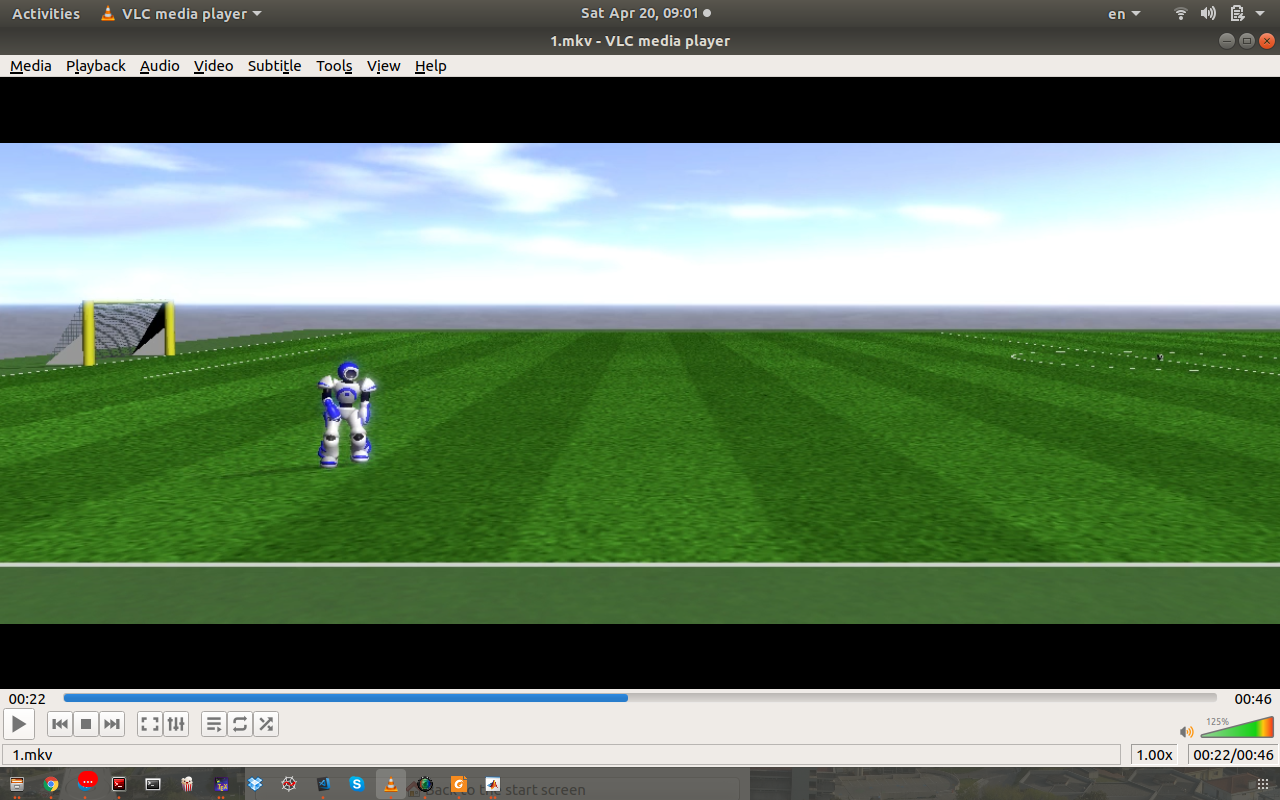}						
	\end{tabular}
	\caption{ Four snapshots of the omnidirectional walking simulation scenario.}
	\label{fig:omni_exp}
	\vspace{-3mm}
\end{figure}

The omnidirectional walking scenario is focused on examining the ability of the proposed walking engine to provide a robust omnidirectional walking. In this simulation, a simulated robot should turn right while walking diagonally (forward and sideward simultaneously) without falling. Indeed, the robot starts from the stop state and should smoothly increase the walking speed to reach its target velocity. It should be noted that in this simulation, the walking parameters have been tuned using some trial and error. Four snapshots of this experiment are shown in figure~\ref{fig:omni_exp}. A video of this simulation is available online at: \url{www.dropbox.com/s/z99xpncxscje97z/OmniWalk.mkv?dl=0}.  

\subsection{Maximum Walking Speed}
One of the major metrics of a walking engine is the maximum velocity of the forward walking. To evaluate this metric, another simulation scenario has been set up. In this simulation, the robot starts from a specific point that is $10~m$ far from the midline of the field. The robot should walk toward this line as fast as possible without deviating to the sides or being unstable. After several simulations, the best hand-tuned velocity that was achieved was $53~cm/s$.
\begin{figure}[!b]
	\centering
	\begin{tabular}	{c c c c }	
	\includegraphics[width=0.23\linewidth, trim= 1cm 2cm 10cm 4.9cm,clip] {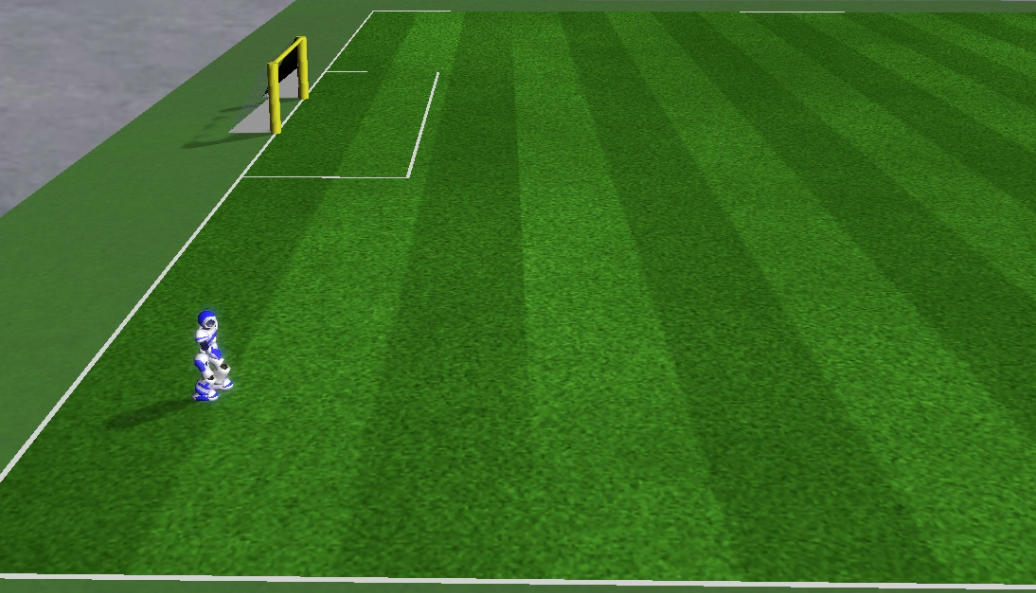}&
	\includegraphics[width=0.23\linewidth, trim= 1cm 2cm 10cm 5cm,clip] {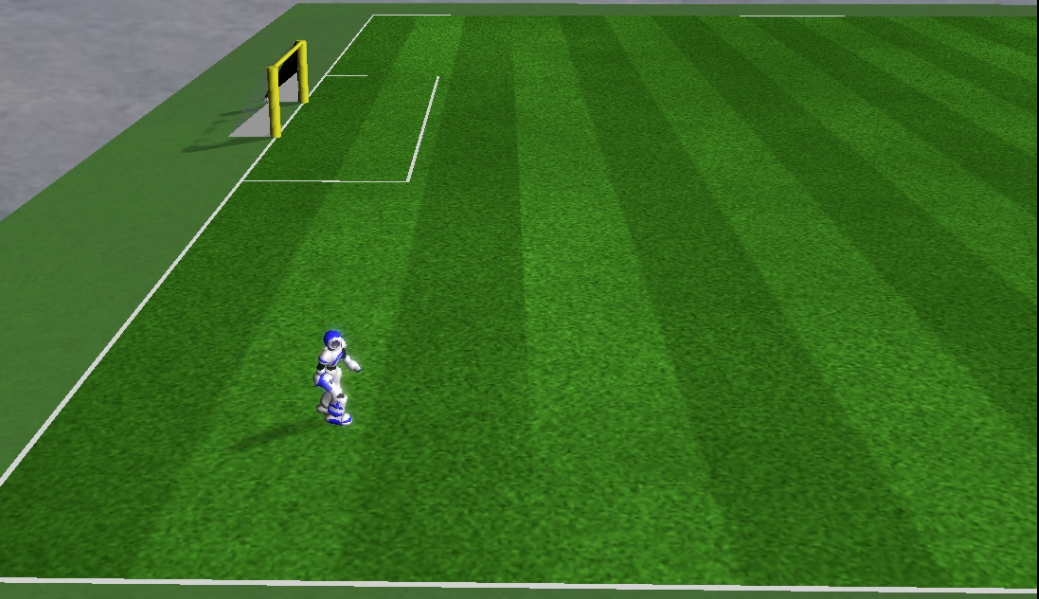}&
	\includegraphics[width=0.23\linewidth, trim= 9cm 2cm 2cm 5.2cm,clip] {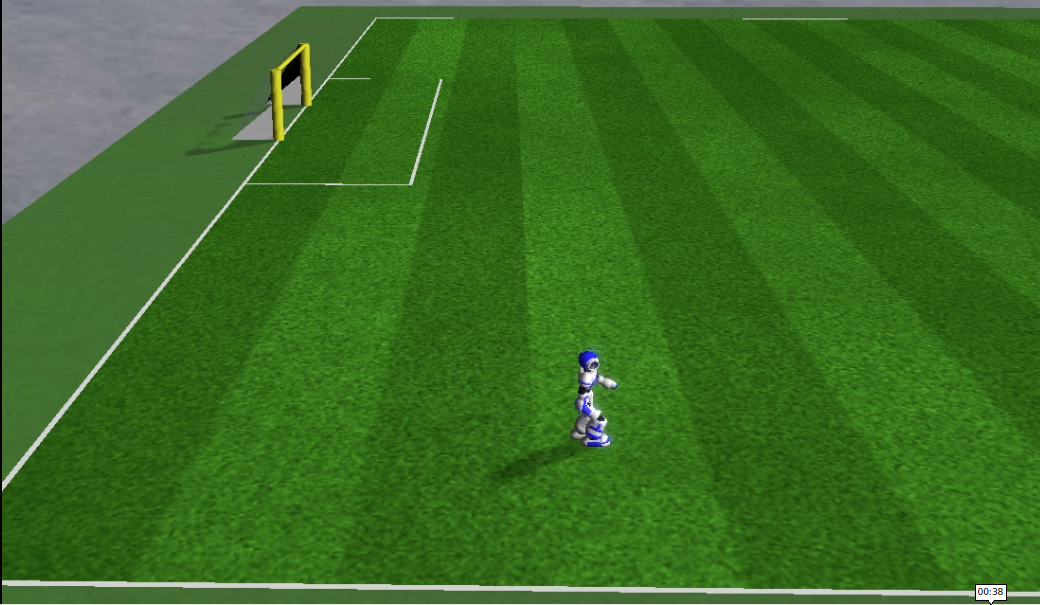}&
	\includegraphics[width=0.23\linewidth, trim= 9cm 2cm 2cm 5.3cm,clip] {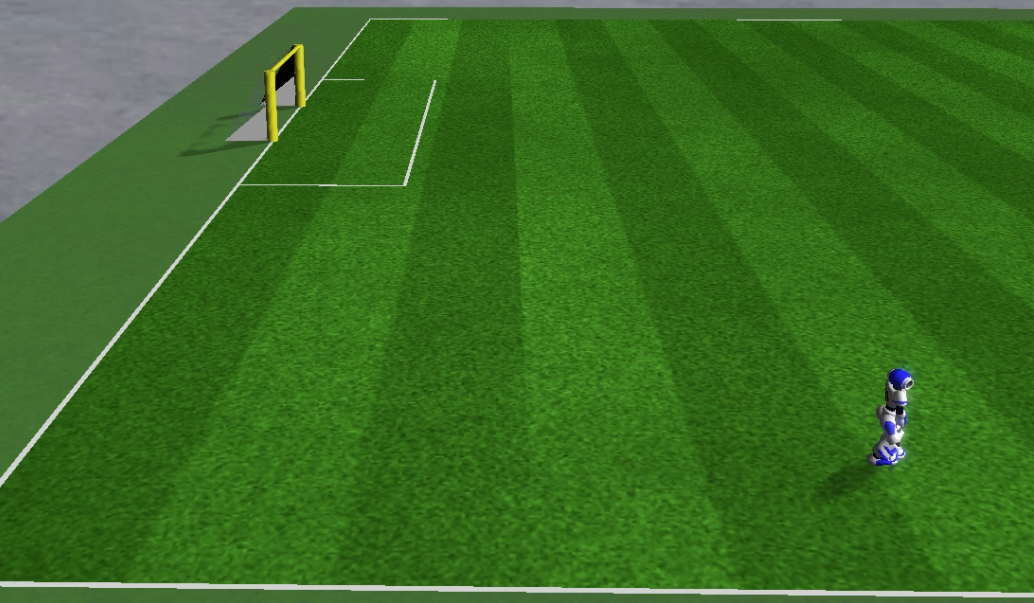}
	\end{tabular}	
	\caption{ Four snapshots of the maximum walking speed simulation scenario.}
	\label{fig:walking_exp}
\end{figure}
As described in previous sections, the proposed framework is fully parametric which allows using optimization algorithms to find the optimum parameter. Therefore, in order to find the maximum forward walking speed, an optimization based on the GA algorithm has been set up to find the optimum values of the parameters. In this optimization, 8 parameters of the framework have been selected to be optimized that were expected step movement~($x$,$y$,$\theta$) , the maximum height of swing leg~($Z_{swing}$), the step duration~($T_{ss}$), the constant torso inclination~{$TI_{to}$}, amplitudes of the center of mass movement~($A_z$) and also torso movement~($A_{to}$). In this scenario, the simulated robot is allowed to walk forward during 10 seconds, and a simple but efficient fitness function $f$ with parameters $\phi$ is defined as:
\begin{equation}
\label{eq:fitness}
 f(\phi)= -|\Delta X| + |\Delta Y| + \epsilon
\end{equation}
\noindent
\begin{wrapfigure}{r}{50mm}
	\vspace{-12mm}
	\center
	\includegraphics[scale=0.16, trim= 6cm 6cm 6cm 6.5cm,clip] {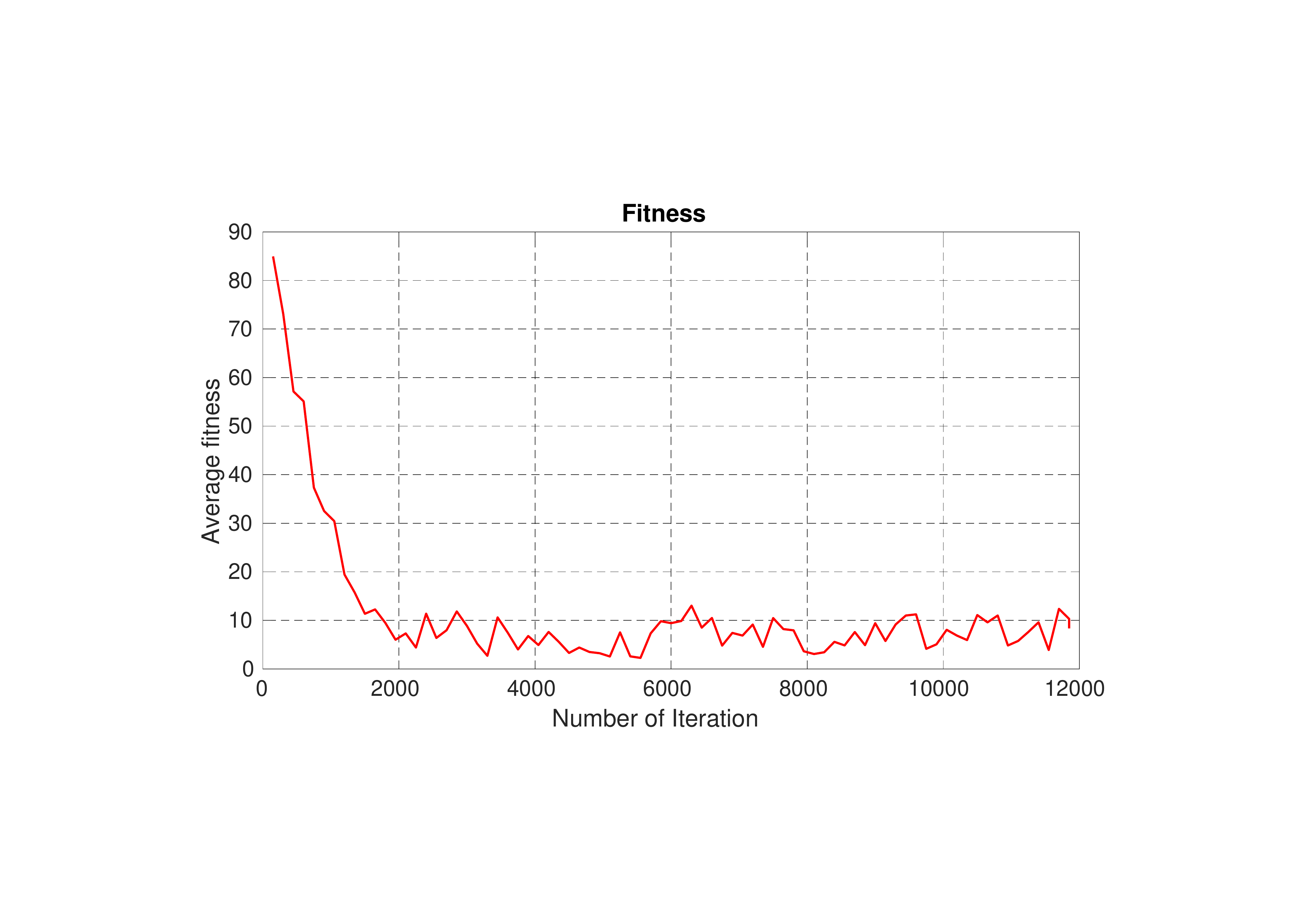}
	\vspace{-3mm}
	\caption{ Evolution of the fitness.}
	\vspace{-8mm}		
	\label{fig:fitness}	
\end{wrapfigure} 
where $\Delta X$ is the total distance covered in~X-axis, $\Delta Y$ represents the  deviated distance in the Y-axis, $\epsilon$ is zero if the simulated robot did not fall and 100 otherwise. Based on this fitness function, the robot is rewarded for moving forward toward the midline of the filed and it is penalized for deviating or falling. Each iteration has been repeated three times and the average of the fitness has been used. After around 12000 iterations of optimization, GA improved the robot walking speed and reached a maximum stable velocity of 80.5~$cm/s$ which is 54\% faster than the best hand-tuned solution. Our optimized walking is faster than the agents in~\cite{asta2011nature},~\cite{kasaei2017hybrid},~\cite{picado2009automatic} and~\cite{shafii2010biped}. The average fitness value of different parameters setting is depicted in figure~\ref{fig:fitness}. Four snapshots of this simulation are shown in figure~\ref{fig:walking_exp}. A video of this simulation is available online at: \url{https://www.dropbox.com/sh/0kk0i0ucoxy7dav/AAAMkDxLlObsmwc2PgL-dl-ha?dl=0}.  

\section{Conclusion}
\label{sec:conclusion}
In this paper, we presented an architecture to generate a model-based walking engine. In particular, we used the ZMP concept as our main stability criterion and extended the LIPM to investigate the effects of vertical motion of COM and also torso inclination while walking. Using the obtained dynamics model, we formulated the problem of the low-level controller of humanoid walking as an optimal controller. We performed some simulations to show how this controller can optimally track the desired trajectories even in the presence of noise. After that, according to the periodic nature of human walking and using the proposed dynamics model and also low-level controller, we modeled the walking engine using a state machine which was composed of four distinct states: Idle, Initialize, Single Support and Double Support. We formulated all the procedures of generating the reference trajectories and validated them using designing an omnidirectional waking simulation scenario. The simulation result shows that our framework is able to generate stable omnidirectional walking. We carried out several simulations and tuned the parameters manually to find the maximum forward walking speed that the simulated robot can walk using our framework. The best hand-tuned speed that we achieved was 53$cm/s$. In order to find the optimum value of the parameters and improve the walking speed, we selected 8 major parameters of our framework and optimized them using GA. After 12000 iterations, the walking speed reached a maximum stable velocity of 80.5~$cm/s$ which was 52\% faster than our best hand-tuned version and also was faster than~\cite{asta2011nature},~\cite{kasaei2017hybrid},~\cite{picado2009automatic} and~\cite{shafii2010biped}.

In future work, we would like to investigate the effects of considering the mass and inertia of the swing leg in the dynamics model. Additionally, we intend to port the proposed framework to the real hardware to show the performance of this framework. 
\vspace{-2mm}
\section*{Acknowledgement}
This research is supported by Portuguese National Funds through Foundation for Science and Technology (FCT) through FCT scholarship SFRH/BD/118438/2016.
\vspace{-2mm}
\bibliographystyle{splncs03}
\bibliography{main}

\end{document}